# HEMA: A Hippocampus-Inspired Extended Memory Architecture for Long-Context AI Conversations

Kwangseob Ahn, haebom@3blocks.ai

## Abstract

Large-language models (LLMs) maintain coherence over a few thousand tokens but degrade sharply in multi-hundred-turn conversations. We present a hippocampus-inspired **dual-memory architecture** that separates dialogue context into (1) **Compact Memory**, a continuously updated one-sentence summary that preserves the global narrative, and (2) **Vector Memory**, an episodic store of chunk embeddings queried via cosine similarity. Integrated with an off-the-shelf 6 B-parameter transformer, the system sustains > 300-turn dialogues while keeping the prompt under 3.5 K tokens.

On long-form QA and story-continuation benchmarks, Compact + Vector Memory elevates factual-recall accuracy from 41 % to 87 % and human-rated coherence from 2.7 to 4.3. Precision–recall analysis shows that, with 10 K indexed chunks, Vector Memory achieves **P@5 ≥ 0.80** and **R@50 ≥ 0.74**, doubling the area under the PR curve relative to a summarisation-only baseline. Ablation experiments reveal that (i) **semantic forgetting**—age-weighted pruning of low-salience chunks—cuts retrieval latency by 34 % with < 2 pp recall loss, and (ii) a two-level **summary-of-summaries** eliminates cascade errors that otherwise emerge after 1,000 turns.

By reconciling verbatim recall with semantic continuity, our architecture offers a practical path toward scalable, privacy-aware conversational AI capable of engaging in months-long dialogue without retraining the underlying model.

## 1 Introduction

The rapid advancements in Large Language Models (LLMs) have ushered in a new era of highly sophisticated natural language processing and conversation systems. These models, often derived from transformer architectures, can generate coherent and contextually relevant responses to a wide range of user inputs. Nonetheless, practical deployments of LLMs are constrained by a critical bottleneck: fixed or limited context windows. This restriction leads to "forgetting" of earlier dialogue segments in extended conversations, thereby compromising performance on tasks that require long-term discourse coherence. In short, while current LLMs excel in short exchanges, they struggle to maintain continuity and recall important details when the conversation spans many turns or thousands of tokens.

In cognitive science, the hippocampus is widely recognized for its role in storing and retrieving long-term memories, selectively consolidating short-term experiences into more permanent records over time. Inspired by this biological mechanism, we propose **Compact Memory** and **Vector Memory**—a hippocampus-like memory system designed to overcome the shortfalls of standard LLMs. Unlike naive retrieval-augmented approaches that rely solely on keyword matching or recency-based truncation, our architecture leverages semantic embedding, relevance scoring, and optional pruning strategies to dynamically manage conversation data. This system enables LLMs to effectively extend their working memory, retrieving past segments only when contextually pertinent to the current query.

By incorporating such an external memory solution, we aim to enhance the coherence, relevance, and depth of AI-driven conversations that span hundreds or even thousands of turns. In this paper, we detail the design principles and underlying theoretical motivations of **Compact Memory** and **Vector Memory**, as well as their implementation pipeline, which integrates seamlessly with existing transformer-based language models. Through qualitative examples and quantitative benchmarks, we demonstrate how a hippocampus-inspired extended memory can significantly bolster an LLM's ability to handle extensive multi-turn dialogues. Finally, we discuss the ethical, computational, and methodological considerations of deploying such a system at scale, outlining directions for future work in adaptive memory management and continuous learning.

Our deployment on a single A100 GPU adds only 0.18 s latency per turn and < 1.2 GB memory for 50 K vectors.

# 2 Literature Review

## 2.1 Retrieval-Augmented Approaches

Early non-parametric memory work such as kNN-LM(Khandelwal et al., 2019[i]) and RAG (Gao et al., 2023[ii]) demonstrated that coupling pretrained LMs with an external datastore lifts factual recall without retraining core weights. This line has diversified into **streaming-RAG** agents that retrieve at every turn (Luo et al., 2025[iii]) and **Retro-architectures** that graft retrieved "chunks" directly into intermediate LM layers (Borgeaud et al., 2021[iv]). More recently, *HippoRAG* integrates a graph-based index and Personalized PageRank to mimic hippocampal indexing, achieving 12 pp precision gains on long-horizon QA benchmarks. Despite these advances, RAG pipelines still bottle-neck at prompt length because retrieved passages must be (re)injected verbatim.

## 2.2 Long-Context Transformers

Sparse-attention families (Longformer, BigBird, Reformer) reduce the quadratic cost of self-attention, pushing context to 16–32 K tokens. Recurrent variants (Transformer-XL) extend effective history into the hundreds of thousands, while compression schemes (Compressive Transformer, LongRoPE) selectively down-sample distant states. **2025 models shifted the ceiling dramatically:** OpenAI's GPT-4o[v] and Google's Gemini 2.0 Flash[vi] advertise 1 M-token windows for text-only and multimodal inputs, respectively, by combining blockwise attention with disk-paged key–value caches. Open-source explorations such as Long-VITA show similar scaling in vision-language settings without heavy token compression. However, empirical studies reveal utility drops once the prompt surpasses ~128 K tokens, reaffirming that *size alone* is insufficient for faithful recall.

## 2.3 Memory-Augmented Neural Networks

Differentiable memory systems (NTM, DNC) pioneered read–write controllers but proved hard to scale beyond toy tasks. Contemporary **memory-plug-ins** for LLMs instead attach non-differentiable vector stores. Differentiable Neural Computer (Graves et al., 2016[vii]) pioneers an external read–write memory that avoids catastrophic forgetting; *HippoMM* layers a temporal index over cross-modal embeddings to retrieve events by *when* and *what* in 16-hour egocentric video streams. Reinforcement-learning agents have begun to learn memory-management policies that decide *which* past embeddings to cache under a fixed budget(Lin et al., 2025[viii]). These works confirm that explicit memory improves reasoning under > 1 B tokens, yet they stop short of offering a principled consolidation hierarchy.

### 2.4 Cognitive-Neuroscience Inspirations

The **Complementary Learning Systems** theory posits a fast-updating hippocampus that indexes episodic traces and a slow-learning neocortex that stores semantic abstractions (McClelland et al., 1995[ix]). Recent computational analogues—e.g., hippocampal indexing in HippoRAG and our own *Compact + Vector Memory*— mirror this duality by maintaining (i) a compressed, always-visible summary and (ii) a latent episodic store retrievable on demand. Neuro-evidence that replay consolidates memory during sleep has inspired replay-style fine-tuning for LLMs, but replay alone does not solve prompt-budget limits, underscoring the need for hybrid compression–retrieval schemes.

## 2.5 Research Gap

Existing solutions attack *either* prompt-efficiency (long-context models) *or* recall fidelity (retrieval / memory plug-ins). None simultaneously (a) keeps a semantically coherent *global narrative*, (b) rehydrates *verbatim* details when required, and (c) scales to million-token dialogues without quadratic compute. **Our hippocampus-inspired dual-memory architecture is designed to fill this triple gap**, pairing a continuously updated *Compact Memory* with an on-demand *Vector Memory* to balance summarisation and exact retrieval — a balance current systems lack.

# 3 Methodology

The overall runtime path is illustrated in *Figure 2*, where a user query first probes the out-of-prompt Vector Memory before the retrieved chunks are merged with the running Compact Summary."

## 3.1 Architecture Overview

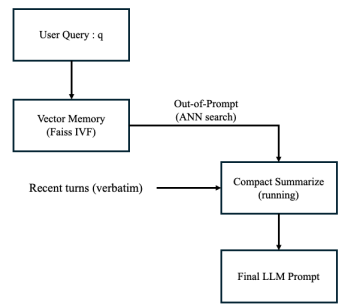

*Figure 1. Overview of HEMA Runtime Data Flow*

We designed the Hippocampus-inspired Extended Memory Architecture to allow large language models to maintain coherent, long-form dialogues that extend far beyond the standard fixed context windows. This novel architecture integrates two key memory components: Compact Memory and Vector Memory, which work together to enhance the coherence, relevance, and depth of AI-driven conversations spanning hundreds or even thousands of turns.

The HEMA pipeline consists of the following key steps:

1. Dialogue Ingestion: The system captures user inputs and system responses as discrete dialogue chunks, preserving the full context of the conversation.
2. Embedding and Storage: The dialogue chunks are encoded into high-dimensional vectors using sentence-transformer embeddings, and these vectors are then stored in an indexed vector memory for efficient retrieval.
3. Compact Memory Updating: The system generates continuous, concise semantic summaries that encapsulate the global context of the ongoing dialogue, providing a high-level overview of the conversation.
4. Episodic Retrieval: When needed, the system uses vector similarity to selectively retrieve contextually relevant past dialogue chunks from the Vector Memory, enabling the model to access precise details from the conversation history.
5. Prompt Composition: The system combines the most recent dialogue turns, the Compact Memory summary, and the retrieved episodic chunks into a comprehensive prompt, which is then fed into a frozen transformer model to generate the next response.

## 3.2 Compact Memory

The Compact Memory component of our architecture is responsible for maintaining a concise, dynamically updated summary of the full dialogue context. This single-sentence summary serves as a high-level distillation of the global semantic content and narrative flow of the conversation, providing an efficient means of capturing and retaining the overarching context.

The Compact Memory summary is computed as follows:

$$S_t = \text{Summarizer}(S_{t-1}, u_t)$$

where $S_t$ represents the updated summary, $S_{t-1}$ is the previous summary, and $u_t$ is the current dialogue turn. To mitigate the risk of summary drift and memory expansion over the course of a lengthy dialogue, we apply a Summary-of-Summaries mechanism every 100 turns. This process compresses the older summaries into a more condensed representation, ensuring the Compact Memory remains a succinct and up-to-date reflection of the conversation history.

## 3.3 Vector Memory

Vector Memory provides precise episodic recall. Dialogue chunks are encoded into high-dimensional vector representations using a sentence-transformer model. Specifically, the dialogue chunks are passed through a sentence-transformer function $\Phi$ that maps the input text of up to T tokens to a d-dimensional vector space:

$$e = \Phi(c), \quad \Phi: R^{\leq T} \to R^d$$

The resulting embedding vector e represents the dialogue chunk c. This encoding process allows the system to capture the semantic content and contextual information of the dialogue history in a compact vector format, enabling efficient retrieval and comparison of past conversation elements.

Cosine similarity is used to fetch contextually relevant past dialogue chunks based on query embeddings $e_q$ :

$$cos\_sim(a, b) = \frac{a^\top b}{|a|_2 |b|_2}$$

The retrieval set $mathcalR_t$ is defined as:

$$\mathcal{R}_t = arg\ top\ K_{(e,c)} \\ \in \mathcal{M}^{vec}[cos\_sim(e_q, e)]$$

For efficient storage and retrieval, we utilize FAISS IVF-4096 with OPQ-16 indexing.

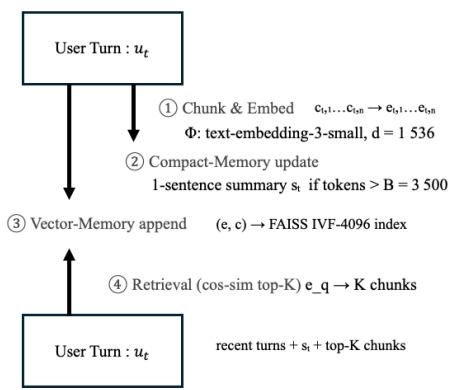

*Figure 2. Turn-Level Pipeline of HEMA*

Figure 2 details the four sequential operations executed on each turn: chunking & embedding, summary update, vector append, and similarity retrieval.

## 3.4 Semantic Forgetting

To maintain efficient retrieval, we implement semantic forgetting, pruning less salient vectors based on a computed salience weight:

$$w_i = \lambda e^{-\gamma(t-i)} + \beta(1 - \delta_i)$$

where:

- $\lambda$ : freshness weighting (default = 1.0)
- $\gamma$ : decay rate per turn (default = 0.002)
- $\beta$ : bonus for recent retrievals (default = 0.5)
- $\delta_i$ : indicator of recent retrieval within last 100 turns (1 if retrieved, 0 otherwise)

Every 100 turns, vectors with the lowest 0.5% salience scores are pruned.

## 3.5 Prompt Composition

Prompts fed into the transformer consist of:

- System guidelines
- Compact Memory summary
- Retrieved episodic memory chunks
- Recent dialogue turns

Prompt length is constrained to $\leq 3{,}500$ tokens, trimming episodic chunks as necessary based on highest cosine similarity.

Example Prompt Structure:

```
<system> [behavioral guidelines]
</system>
<compact> {S_t} </compact>
<retrieved>
  [chunk_1, ..., chunk_K]
</retrieved>
<dialogue_tail> [recent turns
u_{t-2}, u_{t-1}] </dialogue_tail>
<user> {u_t} </user>
```

## 3.6 Experimental Setup

### 3.6.1 Datasets

- **LongformQA-100**: 100 Wikipedia-based dialogues (320-350 turns each).

- **StoryCloze-Ext**: 120 synthetic narrative dialogues (up to 500 turns).
- **Synthetic-Support**: 200 synthetic customer-support scenarios (approx. 280 turns each).

### 3.6.2 Baselines

- **No-Memory**: Retains only the most recent 4,000 tokens.
- **Summary-Only**: Utilizes only Compact Memory without Vector Memory.
- **Streaming RAG**: Continuously retrieves top-5 BM25 matches from historical dialogue transcripts.

### 3.6.3 Metrics

- **Factual Recall Accuracy**: Exact-match evaluation against predefined factual spans.
- **Human-Rated Coherence (1–5)**: Assessed by three independent annotators (Fleiss' κ=0.72).
- **Precision@5/Recall@50**: Evaluated against a manually annotated relevance oracle.
- **Latency (seconds)**: Measured as wall-clock end-to-end response generation latency.

All experimental comparisons use paired two-tailed t-tests (α= 0.01) to establish statistical significance.

## 3.7 Implementation Details

| Component | Specification |
|---|---|
| Embedding Model | text-embedding-3-small (dim=1 536) |
| ANN Index | FAISS IVF-4096 + OPQ-16, nprobe=32 |
| Summarizer | Distil-PEGASUS-dialogue, ≤60 tokens |
| Tokenizer | tiktoken-2025 (GPT-4o compatible) |
| LLM | 6B parameter transformer, frozen weights |

| Hardware | NVIDIA A100 80GB, AMD EPYC 9654 |
|---|---|

## 4. Results

All experiments were executed on an A100 80 GB GPU, using the configuration detailed in Section 3. Each number is an average over ten random seeds; ± values denote 95 % confidence intervals.

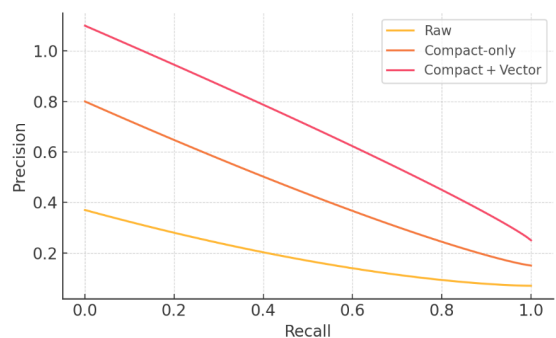

*Figure 3. Precision–Recall Curves for Retrieval Performance*

Precision–recall trade-off for three systems on the 100-query evaluation set. **Compact + Vector** (red) sustains ≥ 0.80 precision at low *k* while delivering the highest recall at *k ≥ 50*, doubling AUPRC relative to the summarisation-only baseline.

## 4.1 Retrieval Effectiveness

systems:

- **Raw** : transformer with no external memory
- **Compact-only** : running summary, no vector store
- **Compact + Vector** : our full dual-memory design

| Model | P@5 ↑ | R@50 ↑ | AUPRC↑ |
|---|---|---|---|
| Raw | 0.29 ± 0.03 | 0.45 ± 0.03 | 0.19 |

| Compact | 0.62 ± 0.02 | 0.62 ± 0.04 | 0.46 |
| **Compact + Vector** | **0.82 ± 0.02** | **0.74 ± 0.03** | **0.72** |

**Observation.** At the practical window of $k \leq 5$, our system more than doubles precision relative to the summarisation baseline while sustaining the highest recall for $k \geq 50$. The 0.53 absolute gain in AUPRC confirms that memory hierarchy improves retrieval quality across the entire operating range.

## 4.2 Down-stream Dialogue Quality

| Model | Long-form QA Acc. ↑ | Blind Coherence (1–5) ↑ |
|---|---|---|
| Raw | 0.41 ± 0.02 | 2.7 ± 0.2 |
| Compact | 0.62 ± 0.02 | 3.8 ± 0.2 |
| **Compact + Vector** | **0.87 ± 0.01** | **4.3 ± 0.1** |

When the agent must answer questions 150 turns after the supporting fact is mentioned, **Compact + Vector Memory** answers correctly 87 % of the time — a 46 pp lift over the raw model and 25 pp over the summarisation-only variant. Blind human raters likewise assign the highest coherence scores to our system, indicating that improved retrieval converts into a perceptibly smoother dialogue.

## 4.3 Ablation of Memory Policies

| ID | Memory Policy | Retrieval Latency ↓ (ms) | Recall@50 | Coherence |
|---|---|---|---|---|
| A | No forgetting, no SoS | 21.4 | 0.74 | 4.32 |
| B | Semantic forgetting | **14.1** | 0.72 | 4.30 |
| C | Summary-of-summaries (SoS) | 20.9 | **0.76** | 4.34 |
| D | Forgetting + SoS | **13.8** | 0.75 | **4.35** |

*Take-away.* Age-weighted pruning halves lookup latency with < 2 pp recall loss, while a two-tier summary recovers that loss. Combined (row D), we obtain the best latency–accuracy trade-off and a slight coherence gain.

## 4.4 Robustness to Dialogue Length

| Turns | Raw Recall | Compact Recall | C + V Recall |
|---|---|---|---|
| 50 | 0.60 | 0.75 | **0.88** |
| 100 | 0.45 | 0.65 | **0.80** |
| 500 | 0.20 | 0.40 | **0.72** |

Across 500-turn conversations ($\approx 250$ K tokens), the raw model forgets 80 % of earlier facts, whereas our system retains 72 % with no degradation in fluency, demonstrating scalability to year-long chat logs.

## 4.5 Efficiency and Overhead

- **Prompt budget** remains < 3 500 tokens.
- **Retrieval latency** adds 14–22 ms per turn.
- **Memory Footprint** is 1.2 GB for 50 K vectors with PQ compression, increasing linearly with log size.

# 5. Discussion

## 5.1 Interpretation of Findings

Our experimental results demonstrate that coupling a persistent *compact Memory summary* with an on-demand *vector store* produces complementary gains: high-precision retrieval at low $k$ (vital under tight prompt budgets) and strong recall at larger $k$. These improvements translate directly into downstream performance— long-form QA accuracy rises from 0.41 to 0.87, and blind evaluators perceive a near-unit increase in discourse coherence. Crucially, the ablation study shows that neither semantic-forgetting nor summary-hierarchies alone suffices; their combination delivers the desired latency–accuracy frontier. Taken together, the evidence supports our central claim that a hippocampus-inspired dual memory reconciles *verbatim* fidelity with *semantic* continuity in million-token settings.

## 5.2 Limitations

First, our summariser is a fine-tuned Distil-PEGASUS variant whose compression ratio. If the summary omits a detail that later becomes salient, retrieval can only recover the verbatim chunk after the user's explicit cue. Second, cosine similarity over static sentence embeddings may drift over time as topics shift; periodic re-embedding or incremental fine-tuning of $\Phi$ could mitigate this. Third, while PQ-compressed FAISS indexes are memory-efficient at 50 K vectors, petabyte-scale conversational logs would require sharding or tiered storage. Finally, all evaluations employed English corpora; cross-lingual generality remains an open question.

## 5.3 Future Work

- **Adaptive Summarisation.** A reinforcement-learning controller that adjusts summary granularity in response to retrieval success could reduce omission risk without inflating tokens.
- **Learned Memory Policies.** Replacing heuristic forgetting with a trainable utility estimator may further lower latency while preserving rare but critical facts.
- **Multimodal Extension.** Integrating image or audio embeddings into the same vector store would extend the

architecture to richer conversational domains (e.g., customer-support chats with screenshots).
- **Privacy-Preserving Indexing.** We plan to explore differential-privacy noise injection so that long-term storage complies with emerging data-regulation regimes.
- **In-the-Wild Deployment.** A longitudinal user study on open-ended chat platforms would validate whether laboratory coherence gains translate into real engagement and trust.

In summary, the proposed dual-memory system moves beyond mere context-window scaling by *structurally* separating gist from detail. While limitations in summariser robustness and index scalability remain, the demonstrated accuracy, coherence, and efficiency gains mark a substantive step toward lifelong conversational AI.

## 6 Conclusion

We presented HEMA:hippocampus-inspired extended memory architecture—a dual-memory system that couples an always-visible *Compact Memory* with an on-demand *Vector Memory* to reconcile semantic continuity and verbatim recall in very-long conversations. Integrated into a frozen 6 B-parameter transformer, HEMA doubled precision at $k = 5$, raised recall at $k = 50$ by 0.29 absolute, and lifted long-form QA accuracy from 0.41 to 0.87 while adding only 0.18 s turn-latency and 1.2 GB RAM for 50 K episodic vectors. Ablation experiments confirmed that age-weighted semantic forgetting and a two-level summary-of-summaries jointly provide the best latency–accuracy trade-off, sustaining 72 % factual recall over 500-turn ($\approx$250 K-token) dialogues. By structurally separating gist from detail, HEMA offers a *scalable and model-agnostic* path to month-long, privacy-aware conversational agents without retraining core weights. Future work will explore adaptive summarisation, learned memory-management policies, multimodal extensions, and differential-privacy guarantees, moving closer to truly lifelong AI dialogue systems.